\documentclass{article}
\usepackage{spconf,amsmath,graphicx,verbatim,amssymb,lineno,makecell}
\usepackage{cuted}
\usepackage{color}

\title{InfiNet: Fully Convolutional Networks for Infant Brain MRI Segmentation}
\name{Shubham Kumar$^{\star}$$^{\dagger}$, Sailesh Conjeti$^{\dagger \mathsection}$, Abhijit Guha Roy$^{\dagger}$$^{\diamond}$, Christian Wachinger$^{\diamond}$ and Nassir Navab$^{\dagger}$$^{\ast}$}
\address{$^{\star}$ Indian Institute of Technology, Roorkee, India \\
$^{\dagger}$Computer Aided Medical Procedures, Technical University of Munich, Germany\\
$^{\mathsection}$German Center for Degenerative Diseases (DZNE), Bonn, Germany\\
$^{\ast}$Computer Aided Medical Procedures, Johns Hopkins University, USA\\
    $^{\diamond}$Artificial Intelligence in Medical Imaging (AI-Med), KJP, LMU Munich, Germany   
    }
\begin{document}
\maketitle
\begin{abstract}
We present a novel, parameter-efficient and practical fully convolutional neural network architecture, termed InfiNet, aimed at voxel-wise semantic segmentation of infant brain MRI images at iso-intense stage, which can be easily extended for other segmentation tasks involving multi-modalities. InfiNet consists of double encoder arms for T1 and T2 input scans that feed into a joint-decoder arm that terminates in the classification layer. The novelty of InfiNet lies in the manner in which the decoder upsamples lower resolution input 
feature map(s) from multiple encoder arms. Specifically, the pooled indices computed in the max-pooling layers of each of the encoder blocks are related to the corresponding decoder block to perform non-linear learning-free upsampling. The sparse maps are concatenated with intermediate encoder representations (skip connections) and convolved with trainable filters to produce dense feature maps. InfiNet is trained end-to-end to optimize for the Generalized Dice Loss, which is well-suited for high class imbalance. InfiNet achieves the whole-volume segmentation in under 50 seconds and we demonstrate competitive performance against multiple state-of-the art deep architectures and their multi-modal variants. 
\end{abstract}
\section{Introduction}
\label{sec:intro}

The first year of the life is the most dynamic stage of the human brain development, which is characterized by rapid tissue growth and development of various cognitive and motor functions~\cite{bcp}.  Non-invasively imaging the infant brain through magnetic resonance imaging (MRI) helps gauge the degree of maturation of the infant's brain. It also aids in assessing risks of the infant developing neuro-developmental and neuropsychiatric disorders in the future~\cite{zhang2015deep}. There is immense clinical advantage to developing tools for automated and objective analysis of the infant brain MRIs, especially for tissue segmentation. This can help the radiologist / pediatrician identify and recognize cues that are not visually apparent and also enables large-scale volumetric studies of the infant brain, such as volume computations, analysis of cortical folding patterns \textit{etc.}~\cite{bcp}. 
In this paper, we particularly focus on infant brain segmentation at the iso-intense stage (6 to 8 months postnatal) using multi-modal MR images (specifically T1 and T2-weighted).

\begin{figure}[t]
\centering
\includegraphics[width=0.45\textwidth]{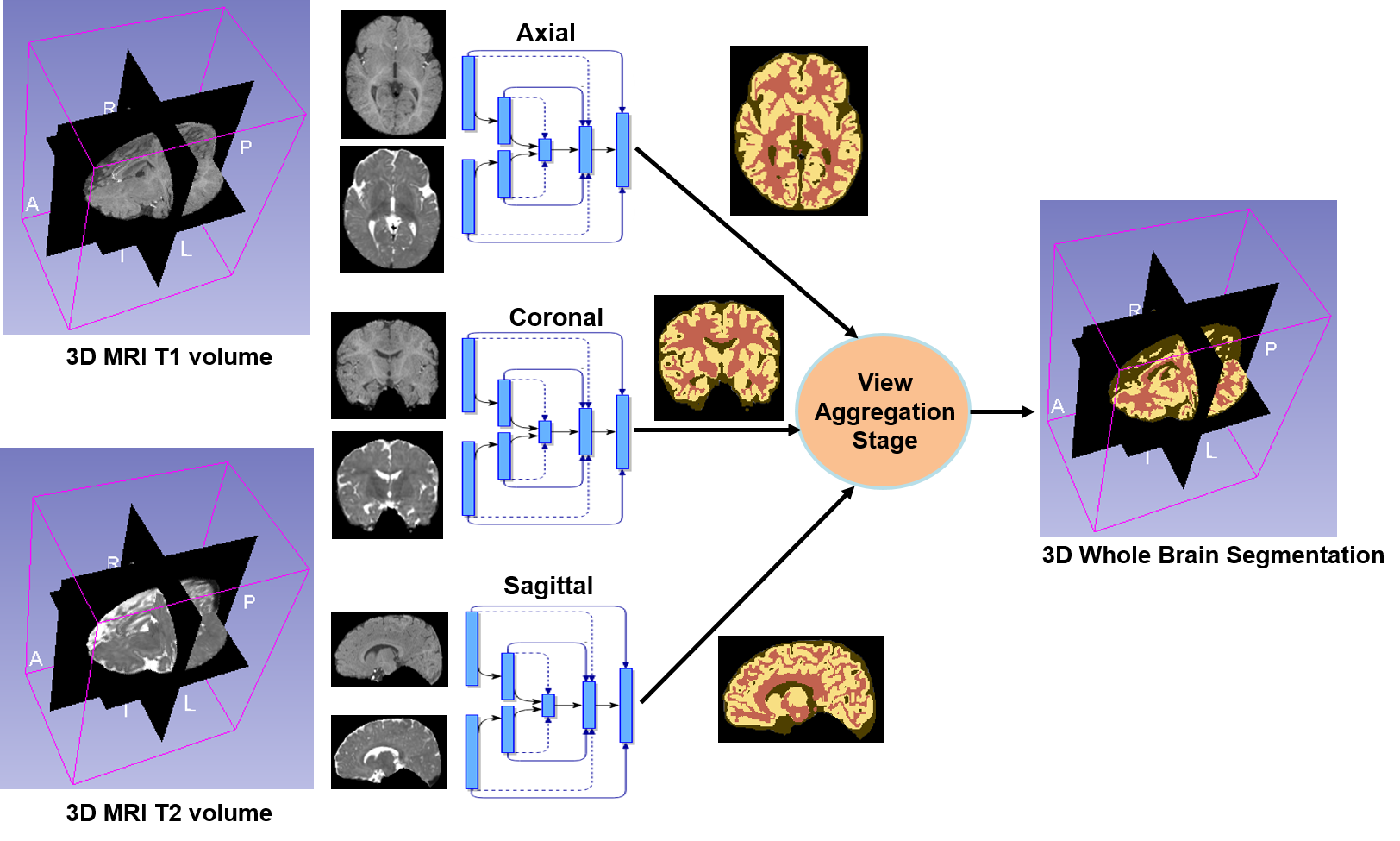}
\caption{Illustration of the overall segmentation, using InfiNet for 3 views followed by view aggregation.}
\label{fig:graphAbstract}
\vspace{-2mm}
\end{figure}

\noindent
\textbf{Prior art}: Segmentation of adult human brain through deep learning has gained a lot of attention 
~\cite{wachinger2017deepnat,ecb}. Their extension to infant brain segmentation is not trivial due to significant anatomical differences between the two age groups. Task-specific approaches in the direction of iso-intense infant brain segmentation include use of handcrafted multi-modal feature fusion approach~\cite{wang2015links}, patch-based multi-modal convolutional neural network (CNN) model ~\cite{zhang2015deep}, to name a few. However, these solutions are either not end-to-end~\cite{wang2015links} or have slow segmentation speed as individual voxels are segmented in sequence. 

\begin{figure*}
  \begin{minipage}[c]{0.6\textwidth}
    \includegraphics[width=\textwidth]{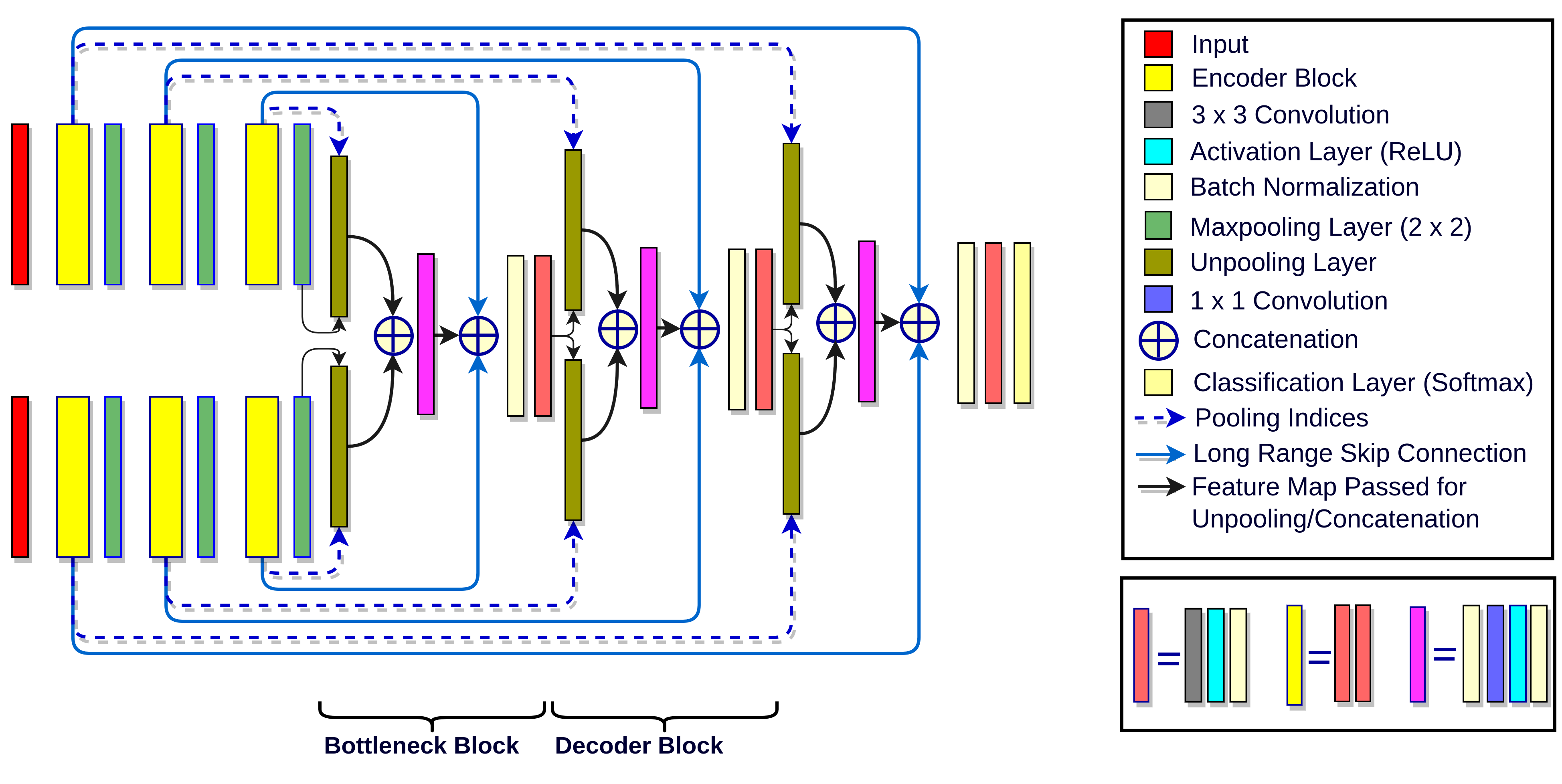}
  \end{minipage}\hfill
  \begin{minipage}[c]{0.3\textwidth}
    \caption{
       Illustration of the proposed InfiNet architecture for segmentation of iso-intense infant brain MRIs. The multiple modalities (T1 and T2) have dedicated encoder arms which feeds into a joint decoder arm for information fusion. The decoder arm terminates in a classification layer to perform slice-wise segmentation.
    } \label{fig:model}
  \end{minipage}
\end{figure*}


\noindent
\textbf{Challenge:} Segmenting this volumes is fraught with several challenges including low Signal-to-Noise ratio (SNR), motion artifacts, poor tissue contrast between the gray and white matter, intensity inhomogeneities \textit{etc.}~\cite{wang2015links}. In addition to these, the high degree of intra-subject variability (due to progressive maturational changes) coupled with enormous inter-subject heterogeneity pose additional challenges for automated segmentation. Learning end-to-end networks for segmentation for the application-at-hand can help learn representations that are robust to these task-specific challenges. As the segmentation is performed with multi-modal data, an effective information fusion scheme is needed.

\noindent
\textbf{Approach:} Recently, Fully convolutional neural networks (F-CNNs) have been effectively used for semantic segmentation both for computer vision~\cite{fcn:long:2015,segnet,deconv} and medical imaging~\cite{unet,vnet,ecb}. These F-CNN models leverage the context of whole image for prediction, and provides labels for all pixels simultaneously, making its deployment very fast.In this paper, we propose a variant of the F-CNN model, termed InfiNet, with two encoder arms to process multi-modal data separately. These extracted features from the two modalities are fused at higher levels of abstraction into a joint decoder arm that terminates in a classification layer to get the final segmentation (shown in Fig.~\ref{fig:model}). We train networks dedicated for segmenting across each anatomical view (coronal, sagittal and axial) and the final volume level segmentation is obtained by aggregating across the multiple views (shown in Fig.~\ref{fig:graphAbstract}).


\section{Methodology}

\subsection{Model Architecture}
The overall architecture of the proposed network is illustrated in Fig.~\ref{fig:model}. It has an encoder-decoder based structure, with two encoder arms for multi-modal inputs (T1 and T2 scans). The two encoder arms merge at the bottleneck block, where high level features and context information from both the arms are fused for decoding. The decoder arm aggregates these features with the high resolution information coming from the lower layers with the help of the long range skip connections from the two encoder arms. Both encoder arms consist of $3$ encoder blocks followed by a max-pool layer, $2$ decoder blocks separated by the bottleneck layer. We discuss the architecture details of the encoder blocks, bottleneck block and decoder blocks as follows.

\noindent
\textbf{Encoder Block:}
Each encoder block consists of a $3\times3$ convolutional layer with 64 output feature maps, followed by a ReLU (Rectifier Linear Unit) activation function and a batch normalization layer. Appropriate padding is provided before every convolution to ensure similar spatial dimensions of input and output. We set the depth and kernel size appropriately so that the effective receptive field covers the entire brain region at the end of encoder arm. It therefore presents a good trade-off between model complexity and effective learning of long range spatial dependencies. Each encoder block is followed by a max pooling layer that reduces the spatial dimension of feature maps by half. The indices corresponding to this max pooling are passed to the associated decoder block for up-sampling, effectively preserving fine-grained spatial information~\cite{deconv}.

\noindent
\textbf{Multi-Modal Decoder Block: }
Each decoder block consists of two unpooling layers, which up-samples the input feature maps, without any additional learnable parameters, in contrast to the up-convolution in U-net~\cite{unet}.  The unpooling layer~\cite{segnet} up-samples the spatial dimension of the input feature map by using the saved indices with maximum activation during max pooling of the corresponding encoder block. The remaining locations are imputed with zeros. Use of unpooling as the upsampling layer reduces the model complexity significantly and at the same time enables the model to generate segmentation maps with finer details. Such a design choice is particularly suited for learning scenarios with limited training data, such as the application at hand.
The aforementioned unpooled feature maps are concatenated into a batch normalization layer, followed by a $1\times1$ convolutional layer to reduce the number of feature maps to $64$. Such a dimensionality reduction is introduced to reduce model complexity, thus avoiding over-fitting. 
Post this, a ReLU activation layer and a batch normalization layer are added. This is again concatenated with encoder feature maps with similar spatial dimension, from both the arms \textit{via} long range skip connections. These skip connections not only provide high contextual information to aid segmentation, but also create a resistance-free path for gradients to flow from deeper regions to shallower regions of the network improving trainability~\cite{unet}. This is followed by a $3\times3$ convolutional layer, a ReLU activation layer and another batch normalization layer. Placement of batch normalization layer at each place prevents internal co-variate shifts and over-fitting during training.

\noindent
\textbf{Bottleneck Block: }
The structure of the Bottleneck Block is akin to the Decoder Blocks with the only difference that it performs early fusion of high-level representations from the individual encoder arms by unpooling them separately using the corresponding indices.

\noindent
\textbf{Classification Layer:}
The classifier consists of a $1\times1$ convolutional layer to transfer the $64$ dimensional feature map to a dimension corresponding to the number of classes. It is followed by a softmax layer and a loss layer. 

\noindent
\textbf{Cost Function:}
The proposed network is learnt by optimizing the Generalized Dice Loss (GDL\textsubscript{v})~\cite{GDL} as the loss function, which is a variant of the Dice Loss. This loss function was chosen to compensate for the inherent class imbalance in the segmentation task. The weights for different classes are determined by the inverse square of their frequencies in the training data. Given the estimated probability at pixel $\mathbf{x}$ to belong to the class $l$ is $p_l(\mathbf{x})$ and the ground truth probability $g_l(\mathbf{x})$, the loss function is given by
\begin{equation}
\text{GDL}_{\text{v}} =
1-2\cdot\frac{  \sum_{\mathbf{x}} \omega_{l}(\mathbf{x}) p_{l}(\mathbf{x}) g_{l}(\mathbf{x})}{\sum_{\mathbf{x}} \omega_{l}(\mathbf{x})(p_{l}(\mathbf{x}) +g_{l}(\mathbf{x}))}.
\label{eq:cost_gdl}
\end{equation}
Where $\omega(\mathbf{x})$ corresponding to each of the classes $l$ is estimated as $\omega_{l}(\mathbf{x}) = \frac{1}{{f_{l}}^2}$.

\subsection {View Aggregation}

In our proposed framework, the segmentation is performed slice-wise. To effectively use cross-view information and improve inter-frame consistency, we trained three independent networks dedicated to segment along three different principal axes \textit{viz.} axial, coronal and sagittal. The generated volume-level probability maps are aggregated through averaging to get the final segmentation. 
This is deemed essential as some cortical folds and sub-cortical structures are well represented in a particular view. Additionally, view aggregation helps in regularizing the prediction for a given voxel by considering the votes from different views. 
Alternatively, a fully 3D convolutional network~\cite{vnet} can be potentially adopted upon availability of larger training datasets. 

\section{Experiments}
\noindent
\textbf{Dataset:}
In our experiments, we used the dataset acquired from the pilot study of Baby Connectome Project (BCP)~\cite{bcp}, which was made public as part of the iSeg MICCAI Grand Challenge. The details of the scanning protocols and resolution are provided in the challenge website \footnote{http://iseg2017.web.unc.edu/}. 
The dataset consists of 23 volumes of co-registered 
(with 1mm isotropic resolution)T1 and T2 scans from 23 subjects with manually annotated grey matter (GM), white matter (WM) and cerebro spinal fluid (CSF) regions, split into 10 volumes for training and rest were held out for testing. In our experiments, we split the data into $8$ subjects for training and $2$ for testing.

\noindent
\textbf{Network Configuration:}
The networks were trained using mini-batch stochastic gradient descent with momentum, until convergence. The learning rate was initially set to $0.01$ and decreased by one order after every $10$ epochs. Training was conducted in a workstation with a Titan Xp GPU, with 12GB RAM. Restricted by the GPU memory, a low batch size of $8$ was set during training. To compensate for the noisy gradients due to the small batch size, a high momentum of $0.95$ was set. 

\noindent
\textbf{Comparative Methods and Baselines:}
The proposed InfiNet is compared against the state-of-the-art F-CNN model U-Net~\cite{unet}. Also, to validate the choice of using two encoder arms to process the bi-modal input data, instead of stacking the data as 2-channels to a single encoder, we established baselines using single encoder variant of InfiNet. Also, for fair comparison, we compared against U-Net with two encoder arms, with skip connections to the decoder arm. Median-frequency  balanced cross-entropy and SGD were used to train this network. Finally, to demonstrate the effectiveness of View-Aggregation stage, we compared InfiNet models' performance against individual axis-specific models. 

\section{Results and Discussion}
\begin{figure}[h]
\centering
\includegraphics[width=0.45\textwidth]{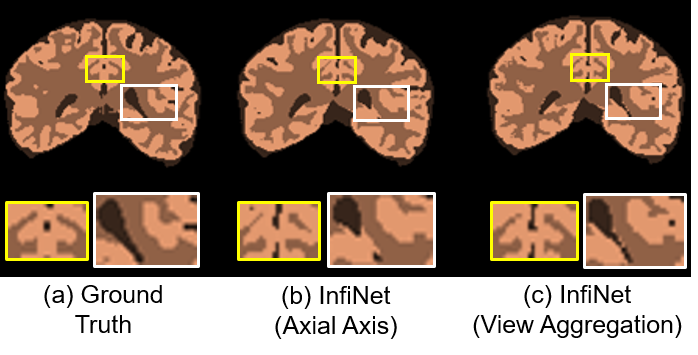}
\caption{Illustration of segmentation using InfiNet with (c) and without (b) view aggregation along with Ground truth (a). Two zoomed regions are highlighted by white and yellow boxes where the difference in segmentation of the cortical folds and CSF is evident.}
\label{fig:Results}
\vspace{-4mm}
\end{figure}

The segmentation performance is evaluated using the Dice score for each of the classes (GM, WM and CSF). The results of InfiNet, its baselines and comparative methods are reported in Tab.~\ref{tab:res}. It is worth noting that InfiNet's score is ~2\% short of the best score in the challenge, which is obtained by the 3D-DenseNet(MSL\_SKKU) \footnote{http://iseg2017.web.unc.edu/rules/results/} which has 1.55 million trainable parameters as compared to just 0.74 million trainable parameters in InfiNet. Comparing single armed U-Net with its double armed variant, we observe an increase of $2\%$, $3\%$ and $3\%$ in dice scores for CSF, GM and WM respectively, substantiating the necessity for processing multi-modal data using two encoder arms and combining the feature at a higher level of abstraction. Comparing double-armed U-Net with InfiNet, trained on coronal slices, we observe an increase of $10\%$, $3\%$ and $3\%$ for CSF, GM and WM respectively, indicating that the choice of Generalized Dice Loss and Unpooling layers is better. This margin is much higher for the CSF class which has a lower class frequency and is tackled effectively using GDL in InfiNet. Among InfiNet variants along the three axes, axial axis provides the best dice score for all the classes. With view aggregation, we achieve the best dice scores which is $2\%$, $3\%$ and $2\%$ above the best axial view InfiNet, substantiating the effect of view aggregation. For the unseen $13$ test subjects in the challenge, we trained ensemble of InfiNets for each view by bootstrapping from the $10$ training volumes and used view aggregation. This resulted in an increase of $2\%$ dice score for all the classes, in comparison to the previous best performance. An example qualitative result is shown in Fig.~\ref{fig:Results} contrasting the performance of InfiNet with and without view aggregation along with ground truth.

 \begin{table}[t]
 \centering
 
\vspace{-4mm}
\caption{Mean Dice scores for the different F-CNN models and training procedures.}
 \begin{tabular}{|c|c|c|c|}
    \hline
     Method & CSF & GM & WM \\
    \hline
    Coronal U-net (Single-Arm) & $0.764$ & $0.784$ & $0.767$ \\
    Coronal U-net (Double-Arm) & $0.779$ & $0.819$ & $0.792$ \\ \hline
    Sagittal InfiNet & $0.912$ & $0.871$ & $0.831$ \\ 
    Coronal InfiNet & $0.898$ & $0.851$ & $0.822$ \\ 
    Axial InfiNet & $0.916$ & $0.875$ & $0.847$ \\ 
    InfiNet (View Aggregated)& $\mathbf{0.926}$ & $\mathbf{0.887}$ & $\mathbf{0.856}$ \\ \hline
    InfiNet (Challenge Test Data) & $\mathbf{0.940}$ & $\mathbf{0.901}$ & $\mathbf{0.880}$ \\ \hline
    3D-DenseNet & $\mathbf{0.958}$ & $\mathbf{0.919}$ & $\mathbf{0.901}$ \\ \hline
  \end{tabular}
  \label{tab:res}
\end{table}

\section{Conclusion}
In this paper, we propose InfiNet, an F-CNN model to process multi-modal inputs using different encoder arms with a joint decoder, followed by a view aggregation stage to leverage information from different views. Using the proposed framework, we addressed the challenging task of infant brain segmentation during iso-intense phase. We optimized the model using Generalized Dice Loss to tackle the issue of class imbalance. We compared our model with state-of-the-art F-CNN model, U-Net and demonstrated better performance. With several baseline variants of InfiNet, we substantiated the effectiveness of using the proposed architecture to process multi-modal data along with view aggregation for medical image segmentation.

\noindent
\textbf{Acknowledgement:}
This work was supported in part by the Faculty of Medicine at LMU (F\"{o}FoLe), research grant from the Centre Digitisation.Bavaria (ZD.B), the NVIDIA corporation and DAAD (German Academic Exchange Service).

\bibliographystyle{IEEEbib}
\bibliography{refs}

\begin{thebibliography}{10}

\bibitem{bcp}
L.~Wang, F.~Shi, P.~Yap, W.~Lin, J.~Gilmore, and D.~Shen,
\newblock ``Longitudinally guided level sets for consistent tissue segmentation
  of neonates,''
\newblock {\em HBM}, vol. 34, no. 7, pp. 1747--1747, 2013.

\bibitem{zhang2015deep}
W.~Zhang, R.~Li, H.~Deng, L.~Wang, W.~Lin, S.~Ji, and D.~Shen,
\newblock ``Deep convolutional neural networks for multi-modality isointense
  infant brain image segmentation,''
\newblock {\em NeuroImage}, vol. 108, pp. 214--224, 2015.

\bibitem{wachinger2017deepnat}
C.~Wachinger, M.~Reuter, and T.~Klein,
\newblock ``Deepnat: Deep convolutional neural network for segmenting
  neuroanatomy,''
\newblock {\em NeuroImage}, 2017.

\bibitem{ecb}
A.~Guha~Roy, S.~Conjeti, D.~Sheet, A.~Katouzian, N.~Navab, and C.~Wachinger,
\newblock ``Error corrective boosting for learning fully convolutional networks
  with limited data,''
\newblock {\em MICCAI}, 2017.

\bibitem{wang2015links}
L.~Wang, Y.~Gao, F.~Shi, G.~Li, J.~Gilmore, W.~Lin, and D.~Shen,
\newblock ``Links: Learning-based multi-source integration framework for
  segmentation of infant brain images,''
\newblock {\em NeuroImage}, vol. 108, pp. 160--172, 2015.

\bibitem{fcn:long:2015}
J.~Long, E.~Shelhamer, and T.~Darrell,
\newblock ``Fully convolutional networks for semantic segmentation,''
\newblock in {\em CVPR}, 2015, pp. 3431--3440.

\bibitem{segnet}
V.~Badrinarayanan, A.~Kendall, and R.~Cipolla,
\newblock ``Segnet: A deep convolutional encoder-decoder architecture for image
  segmentation,''
\newblock {\em TPAMI}, 2015.

\bibitem{deconv}
H.~Noh, S.~Hong, and B.~Han,
\newblock ``Learning deconvolution network for semantic segmentation,''
\newblock in {\em ICCV}, 2015, pp. 1520--1528.

\bibitem{unet}
O.~Ronneberger, P.~Fischer, and T.~Brox,
\newblock ``U-net: Convolutional networks for biomedical image segmentation,''
\newblock in {\em MICCAI}. Springer, 2015, pp. 234--241.

\bibitem{vnet}
F.~Milletari, N.~Navab, and S.A. Ahmadi,
\newblock ``V-net: Fully convolutional neural networks for volumetric medical
  image segmentation,''
\newblock in {\em 3DV}, 2016, pp. 565--571.

\bibitem{GDL}
C.~H Sudre, W.~Li, T.~Vercauteren, S.~Ourselin, and M.~J. Cardoso,
\newblock ``{Generalised Dice overlap as a deep learning loss function for
  highly unbalanced segmentations},''
\newblock {\em ArXiv e-prints}, 2017.

\end{thebibliography}

\end{document}